\documentclass{article}

\usepackage[final, nonatbib]{neurips_2023}

\usepackage[utf8]{inputenc} %
\usepackage[T1]{fontenc}    %
\usepackage{hyperref}       %
\usepackage{url}            %
\usepackage{booktabs}       %
\usepackage{amsfonts}       %
\usepackage{nicefrac}       %
\usepackage{microtype}      %
\usepackage{xcolor}         %

\usepackage{color}
\usepackage{amsmath}
\usepackage{amssymb}
\usepackage{amsthm}
\usepackage{mathtools}
\usepackage{thmtools}
\usepackage[mathscr]{eucal}
\usepackage{bm}
\usepackage{bbm}
\usepackage{relsize}
\usepackage{graphics}
\usepackage{wrapfig}
\usepackage{multirow}
\usepackage{enumitem} 

\usepackage[algo2e]{algorithm2e}

\usepackage[capitalize,noabbrev]{cleveref}

\usepackage[numbers]{natbib}

\usepackage{array}

\usepackage{pgf}
\usepackage{xinttools}
\usepackage{tikz}
\usetikzlibrary{arrows.meta}
\usepackage{textcomp}
\usetikzlibrary{calc,shapes,arrows,positioning,automata,trees}
\usepackage{placeins} 
\usepackage{tabularx}
\usepackage{graphicx}
\usepackage{subcaption} 
\usepackage{listings}
\usepackage{xargs,lipsum,caption,changepage,ifthen}

\usepackage{tikz} 

\usepackage{footmisc}

\usepackage{bigdelim}
\usepackage{colortbl} 
\definecolor{mColor1}{rgb}{0.95,0.95,0.95}
\newcolumntype{a}{>{\columncolor{mColor1}}c}

\usepackage[toc,page]{appendix}

\usepackage{tcolorbox}

\definecolor{solarized@base03}{HTML}{002B36}
\definecolor{solarized@base02}{HTML}{073642}
\definecolor{solarized@base01}{HTML}{586e75}
\definecolor{solarized@base00}{HTML}{657b83}
\definecolor{solarized@base0}{HTML}{839496}
\definecolor{solarized@base1}{HTML}{93a1a1}
\definecolor{solarized@base2}{HTML}{EEE8D5}
\definecolor{solarized@base3}{HTML}{FDF6E3}
\definecolor{solarized@yellow}{HTML}{B58900}
\definecolor{solarized@orange}{HTML}{CB4B16}
\definecolor{solarized@red}{HTML}{DC322F}
\definecolor{solarized@magenta}{HTML}{D33682}
\definecolor{solarized@blue}{HTML}{268BD2}
\definecolor{solarized@cyan}{HTML}{2AA198}
\definecolor{solarized@green}{HTML}{859900}

\newtcolorbox{importantresult}{colback=solarized@yellow!5!white,
colframe=solarized@yellow,parbox, left=0.5mm, right=0.5mm,top=0.5mm,bottom=0.5mm}

\newtcolorbox{importantresult_noparbox}{colback=solarized@yellow!5!white,
colframe=solarized@yellow,parbox=false, left=0.5mm, right=0.5mm,top=0.5mm,bottom=0.5mm}

\newtheorem*{theorem*}{Theorem}
\newtheorem*{definition*}{Definition}

\newcommand\Bw{\bm{w}}
\newcommand\Bx{\bm{x}}
\newcommand\By{\bm{y}}

\newcommand{\dE}{\mathbb{E}}

 \newcommand{\rH}{\mathrm{H}}
\newcommand{\rI}{\mathrm{I}}  
\newcommand{\rK}{\mathrm{K}} 
\newcommand{\rM}{\mathrm{M}}

 \newcommand{\cD}{\mathcal{D}}

\newcommand{\cU}{\mathcal{U}}

\DeclarePairedDelimiterX{\kldiv}[2]{(}{)}{%
  #1\;\delimsize\|\;#2%
}

\DeclarePairedDelimiterX{\mi}[2]{(}{)}{%
  #1\;\delimsize ; \;#2%
}

\DeclarePairedDelimiterX{\di}[2]{(}{)}{%
  #1\;\delimsize ; \;#2%
}

\DeclarePairedDelimiterX{\ce}[2]{(}{)}{%
  #1\;\delimsize ; \;#2%
}

\DeclarePairedDelimiterXPP{\mii}[3]%
   {_{\mathrm{#1}}}{(}{)}{}{#2\;\delimsize ; \;#3%
}

\makeatletter
\newcommand{\dlmf}[1]{%
\citep[%
  \def\nextitem{\def\nextitem{, }}%
  \@for \el:=#1\do{\nextitem\href{http://dlmf.nist.gov/\el}{(\el)}}%
]{Olver:10}%
}
\makeatother

\usepackage{pdflscape} 

\newcolumntype{R}[1]{>{\raggedright\arraybackslash}p{#1}}
\newcolumntype{C}[1]{>{\centering\arraybackslash}p{#1}}
\newcolumntype{L}[1]{>{\raggedleft\arraybackslash}p{#1}}

\usepackage{bigdelim}
\usepackage{colortbl} 
\definecolor{mColor1}{rgb}{0.95,0.95,0.95}

\hypersetup{
   colorlinks=true,
   linkcolor=[RGB]{30, 30, 180},
   citecolor=[RGB]{30, 30, 180},
   urlcolor=magenta,
   pdfborder=0 0 0,
   pdftitle={},
   pdfsubject={}, 
   pdfkeywords={},
   pdfauthor={},%
   pdfstartview=FitH
}

\title{Introducing an Improved Information-Theoretic Measure of Predictive Uncertainty}

\author{%
  Kajetan Schweighofer$^*$ \,\; Lukas Aichberger$^*$ \,\; Mykyta Ielanskyi \; Sepp Hochreiter \\
  \\
  ~ELLIS Unit Linz and LIT AI Lab, Institute for Machine Learning, \\ 
                  ~~Johannes Kepler University Linz, Austria \\ 
  $^*$~Joint first authors
}

\begin{document}

\maketitle

\begin{abstract}
    Applying a machine learning model for decision-making in the real world
    requires to distinguish what the model knows from what it does not.
    A critical factor in assessing the knowledge of a model is 
    to quantify its predictive uncertainty.
    Predictive uncertainty is commonly measured by the entropy
    of the Bayesian model average (BMA) predictive distribution.
    Yet, the properness of this current measure of 
    predictive uncertainty was recently questioned.
    We provide new insights regarding those limitations.
    Our analyses show that the current measure erroneously assumes that the 
    BMA predictive distribution is equivalent to the predictive distribution 
    of the true model that generated the dataset.
    Consequently, we introduce a theoretically grounded measure to overcome these limitations.
    We experimentally verify the benefits of our introduced measure of predictive uncertainty.
    We find that our introduced measure behaves more reasonably in controlled synthetic tasks.
    Moreover, our evaluations on ImageNet demonstrate that our introduced measure is advantageous in real-world applications utilizing predictive uncertainty.
\end{abstract}

\section{Introduction} \label{sec:introduction}

Decision-making with machine learning models in the real world requires
risk assessment based on the predictive uncertainty of the model to be actionable \citep{Apostolakis:90}.
It is essential to avoid models making high-risk, uncertain decisions.
Instead, such decisions should be deferred to human experts or default to a potentially sub-optimal but safe decision.
Therefore, it is key to use grounded measures of predictive uncertainty
and provide estimates for them when deploying machine learning models for decision-making in
the real world \citep{Helton:93, Kendall:17}.

Predictive uncertainty is often categorized into two types, aleatoric and epistemic, 
according to the source of uncertainty \citep{Huellermeier:21}.
We focus on measuring the predictive uncertainty by 
characterizing the predictive distribution $p(\By \mid \Bx)$ 
of outcomes $\By$ for inputs $\Bx$.
Here, \emph{aleatoric} uncertainty refers to the inherent stochasticity of
sampling outcomes from the predictive distribution, 
thus is irreducible.
Further, \emph{epistemic} uncertainty refers to the lack of knowledge 
about the true predictive distribution,
thus is reducible.
Commonly, those uncertainties are assumed to be additive, summing up
to a \emph{total} predictive uncertainty.

{The currently prevalent measure of predictive uncertainty is the entropy of the Bayesian model average (BMA) predictive
distribution, where the epistemic component is given by the \emph{mutual information} $\rI$ \citep{Houlsby:11, Gal:16thesis, Depeweg:18, Huellermeier:21, Mukhoti:23}. 
Recently, concerns about the properness of this current measure have been raised \citep{Wimmer:23}.
By further examining the current measure, we discern that the core issue 
lies in the assumption that the BMA predictive distribution is equivalent to 
the predictive distribution of the true model. 
Therefore, we propose to use a different 
information-theoretic measure that does not build upon the BMA, 
where the epistemic component is given by the \emph{expected pairwise KL-divergence} $\rK$ \citep{Malinin:19}. \parfillskip=0pt\par}

\begin{figure} %
\begin{center}
\includegraphics[width=\textwidth]{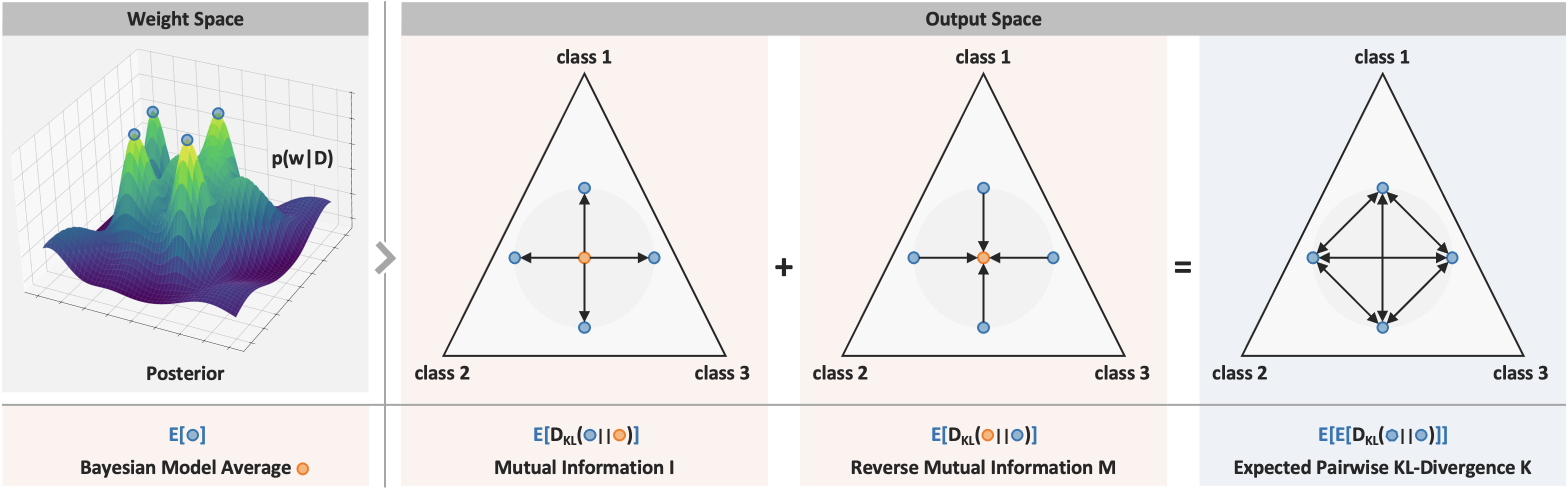}
\caption[]{Relationship between epistemic components of the current and our introduced measure.}
\label{fig:unc_measures}
\end{center}
\vspace{-0.4cm}
\end{figure}

$\rK$ is an upper bound of $\rI$, with the difference being termed \emph{reverse mutual 
information} $\rM$ \citep{Malinin:21} (Fig.~\ref{fig:unc_measures}).
The aleatoric component is the same for both measures.
Our main contributions are as follows:
\begin{itemize}[leftmargin=*, topsep=0pt, itemsep=0pt, partopsep=0pt, parsep=2pt]
    \item We provide new insights regarding the limitations of the current measure of predictive uncertainty.
    \item We introduce a theoretically grounded measure that addresses those limitations (Eq.~\eqref{eq:expected_uncertainty}).
    \item We investigate the empirical benefits of using the theoretically grounded measure.
\end{itemize}

\section{Analyzing the Current Measure of Predictive Uncertainty} \label{sec:old_measures}

Many machine learning models used for classification and regression yield 
the distribution parameters
of a predictive distribution as their outputs. 
The softmax outputs of a classifier define a categorical distribution 
and the outputs of a regressor correspond to the mean (and variance, see e.g. \citep{Kendall:17, Lakshminarayanan:17}) 
of a continuous (e.g. Gaussian) distribution.
A machine learning model with model parameters $\Bw$ is used to obtain estimates of
the distribution parameters of the true predictive distribution
$p(\By \mid \Bx)$ that generated the data.
Therefore, the predictive distribution 
under the model is denoted as $p(\By \mid \Bx, \Bw)$.
We assume the dataset $\cD = \{\Bx_i, \By_i\}_{i=1}^N$ is given, 
thus do not consider uncertainty due to how the dataset was sampled from $p(\Bx, \By)$.
Furthermore, we assume that the true model that created the dataset can be
represented by the chosen model class.
Those are common and often necessary assumptions \citep{Huellermeier:21}.
From a Bayesian point of view, we can assign a posterior probability 
$p(\Bw \mid \cD) \propto p(\cD \mid \Bw) p(\Bw)$ 
of how probable it is that certain model parameters $\Bw$ are the true model parameters $\Bw^*$.
For the true model parameters, $p(\By \mid \Bx, \Bw^*) = p(\By \mid \Bx)$.
The posterior distribution allows marginalization, 
yielding the BMA predictive distribution 
$p(\By \mid \Bx, \cD) = \dE_{p(\Bw \mid \cD)} \left[ p(\By \mid \Bx, \Bw) \right]$.
This expectation is intractable and generally approximated by Monte Carlo sampling of 
model parameters using approximate inference techniques 
\citep{MacKay:92, Neal:95thesis, Welling:11, Blundell:15, Gal:16, Lakshminarayanan:17, Wilson:20}.

The most common measure of predictive uncertainty is the Shannon-entropy $\rH(\cdot)$ \citep{Shannon:48} of the BMA predictive distribution \citep{Houlsby:11, Gal:16thesis, Smith:18, Depeweg:18, Huellermeier:21, Mukhoti:23},
decomposing into aleatoric and epistemic components:
\begin{align}  \label{eq:uncertainty_mi_1}
    \underbrace{\rH ( p(\By \mid \Bx, \cD ) )}_{\text{total}} \ = \ \underbrace{\dE_{p(\Bw \mid \cD )} \left[ \rH  (p(\By \mid \Bx, \Bw ) ) \right]}_{\text{aleatoric}} \ + \ \underbrace{\rI(p(\By, \Bw \mid \Bx, \cD))}_{\text{epistemic}} \ .
\end{align}
The aleatoric component best estimates the true aleatoric uncertainty $\rH (p(\By \mid \Bx, \Bw^*))$, as 
it accounts for all possible models according to their posterior probability.
The epistemic component is the \emph{mutual information} $\rI(p(\By, \Bw \mid \Bx, \cD))$, 
which measures the
reduction of uncertainty about outcomes $\By$ through observing model parameters $\Bw$.
Recently, the properness of the measure given by Eq.~\eqref{eq:uncertainty_mi_1} has been questioned \citep{Wimmer:23}.
Rewriting Eq.~\eqref{eq:uncertainty_mi_1} reveals 
(detailed steps given in Sec.~\ref{sec:equivalence_mi} in the appendix), 
that mutual information is equal to the expected KL-divergence 
$\mathrm{D}_{\mathrm{KL}}(\cdot \mid\mid \cdot)$ between the predictive distributions
of possibly selected models and the BMA \citep{Schweighofer:23, Wimmer:23}.
Furthermore, the entropy of the BMA predictive distribution is equal to the expected
cross-entropy $\mathrm{CE}(\cdot , \cdot)$ between the predictive distributions
of possibly selected models and the BMA:
\begin{align} \label{eq:uncertainty_mi_2}
    &\underbrace{\dE_{p(\Bw \mid \cD)} \left[ \mathrm{CE} (p(\By \mid \Bx, \Bw ) \ , \ p(\By \mid \Bx, \cD )) \right]}_{\text{total}} \\ \nonumber 
    & \qquad\qquad = \ \underbrace{\dE_{p(\Bw \mid \cD )} \left[ \rH  (p(\By \mid \Bx, \Bw ) ) \right]}_{\text{aleatoric}} \ + \ \underbrace{\dE_{p(\Bw \mid \cD )} \left[ \mathrm{D}_{\mathrm{KL}}(p(\By \mid \Bx, \Bw ) \mid\mid p(\By \mid \Bx, \cD )) \right]}_{\text{epistemic}}
\end{align}

Generally, the KL-divergence between two distributions 
$\mathrm{D}_{\mathrm{KL}}(p \mid\mid p^*)$ quantifies the additional surprisal 
when sampling according to $p^*$ instead of $p$ \citep{Cover:06}.
In this setting, the KL-divergence should quantify the epistemic uncertainty, resulting from sampling according to some model's predictive distribution 
$p(\By \mid \Bx, \Bw )$ instead of the true model's predictive distribution
$p(\By \mid \Bx, \Bw^* )$.
However, the epistemic component in Eq.~\eqref{eq:uncertainty_mi_2} uses the
BMA predictive distribution as a surrogate for the 
predictive distribution of the true model.
Those do not coincide in general! 
From a Bayesian point of view, any model parameters could be the true model parameters 
according to their posterior probability.
Furthermore, there could be no model that has nonzero posterior probability 
and equivalent predictive distribution to the BMA.

\section{Introducing a Theoretically Grounded Measure of Predictive Uncertainty} \label{sec:new_measures}

We introduce a theoretically grounded measure of predictive uncertainty
that does not assume that the BMA predictive distribution 
is equivalent to the true model's predictive distribution.
To assess predictive uncertainty, we want to characterize the stochasticity
of sampling outcomes from the predictive distribution $p(\By \mid \Bx)$ that generated the dataset.  
The most sensible information-theoretic approach is the entropy of the
predictive distribution $\rH (p(\By \mid \Bx))$,
capturing the true aleatoric uncertainty \citep{Huellermeier:21}.
As stated in Sec.~\ref{sec:old_measures}, the distribution parameters
of the predictive distribution are usually estimated using a model
with model parameters $\Bw$.
Therefore, an estimate for the aleatoric uncertainty under a given model
is $\rH (p(\By \mid \Bx, \Bw))$.

This approximation gives rise to epistemic uncertainty, as the selected model parameters
$\Bw$ are generally not the true model parameters $\Bw^*$.
We want to measure the additional uncertainty due to the mismatch of
predictive distributions under $\Bw$ and $\Bw^*$.
This can be done using the KL-divergence 
$\mathrm{D}_{\mathrm{KL}} ( p(\By \mid \Bx, \Bw) \mid\mid p(\By \mid \Bx, \Bw^*) )$,
as elaborated for the epistemic component of Eq.~\eqref{eq:uncertainty_mi_2}.
Unfortunately, we don't know the true model parameters $\Bw^*$ in general.
However, the posterior distribution $p(\Bw \mid \cD)$ expresses how likely
certain model parameters are the true model parameters.
Therefore, we can perform an expectation over the model posterior: 
$\dE_{p(\tilde\Bw \mid \cD)} \left[\mathrm{D}_{\mathrm{KL}} ( p(\By \mid \Bx, \Bw) \mid\mid p(\By \mid \Bx, \tilde{\Bw}) ) \right]$.
This represents the expected mismatch between the predictive distribution 
under certain model parameters and all potential parameters, 
weighted by how likely they are the true parameters.
Therefore, the predictive uncertainty of a specific model $\Bw$
is given by
\begin{align} \label{eq:pre_selected_uncertainty}
    & \underbrace{\dE_{p(\tilde\Bw \mid \cD)} \left[ \mathrm{CE} ( p(\By \mid \Bx, \Bw) \ , \ p(\By \mid \Bx, \tilde\Bw) ) \right]}_{\text{total}} \\ \nonumber
    & \qquad\qquad = \ \underbrace{\rH  (p(\By \mid \Bx, \Bw ) )}_{\text{aleatoric}} \ + \     \underbrace{\dE_{p(\tilde\Bw \mid \cD)} \left[ \mathrm{D}_{\mathrm{KL}} ( p(\By \mid \Bx, \Bw) \mid\mid p(\By \mid \Bx, \tilde\Bw) ) \right]}_{\text{epistemic}} \ .
\end{align}
This measure was introduced in Schweighofer et al. 
\citep{Schweighofer:23} to quantify the 
predictive uncertainty of a given, pre-selected model.
This is a critical aspect when deploying a machine learning model in real-world applications.
However, it expresses a subjective uncertainty about the prediction of a specific model.
In contrast, the measure given by Eq.~\eqref{eq:uncertainty_mi_2} expresses
the expected uncertainty when selecting a model according to the posterior.
Thus, we propose to take a posterior expectation of Eq.~\eqref{eq:pre_selected_uncertainty},
resulting in a measure of predictive uncertainty that does not share the limitations of Eq.~\eqref{eq:uncertainty_mi_2}:
\begin{align} \label{eq:expected_uncertainty}
    & \underbrace{\dE_{p(\Bw \mid \cD)} \left[ \dE_{p(\tilde\Bw \mid \cD)} \left[ \mathrm{CE} ( p(\By \mid \Bx, \Bw) \ , \ p(\By \mid \Bx, \tilde\Bw) ) \right] \right]}_{\text{total}} \\ \nonumber
    & \qquad = \ \underbrace{\dE_{p(\Bw \mid \cD)} \left[\rH  (p(\By \mid \Bx, \Bw ) ) \right]}_{\text{aleatoric}} \ + \     \underbrace{\dE_{p(\Bw \mid \cD)} \left[\dE_{p(\tilde\Bw \mid \cD)} \left[ \mathrm{D}_{\mathrm{KL}} ( p(\By \mid \Bx, \Bw) \mid\mid p(\By \mid \Bx, \tilde\Bw) ) \right] \right]}_{\text{epistemic}} \ .
\end{align}
The aleatoric component of Eq.~\eqref{eq:expected_uncertainty} and  Eq.~\eqref{eq:uncertainty_mi_2} are the same.
What differs is the epistemic component, where in Eq.~\eqref{eq:expected_uncertainty} 
it is a pairwise comparison between the predictive distributions of 
possible models and weights according to their posterior probability.
The epistemic component of Eq.~\eqref{eq:expected_uncertainty} was discussed by 
Malinin et al. \citep{Malinin:19} as a measure of ensemble diversity,
called the \emph{expected pairwise KL-divergence} $\rK(p(\By, \Bw \mid \Bx, \cD))$.
However, in follow-up work, Malinin et al. \citep{Malinin:21} erroneously concluded that
solely the mutual information 'cleanly' decomposes into total and aleatoric uncertainty.

\textbf{Relation between measures. }
The expected pairwise KL-divergence is an upper bound of the
mutual information by Jensen's inequality.
The difference between those two is called the
\emph{reverse mutual information} \citep{Malinin:21}, defined as
$\rM(p(\By, \Bw \mid \Bx, \cD)) = \dE_{p(\tilde{\Bw} \mid \cD )} \left[ \mathrm{D}_{\mathrm{KL}}(p(\By \mid \Bx, \cD ) \mid\mid p(\By \mid \Bx, \tilde{\Bw} ) ) \right]$.
Furthermore, the expected pairwise KL-divergence, the mutual information
and the reverse mutual information satisfy 
$\rK(p(\By, \Bw \mid \Bx, \cD))  =  \rI(p(\By, \Bw \mid \Bx, \cD))  +  \rM(p(\By, \Bw \mid \Bx, \cD)) \ $ \citep{Malinin:21}.
A proof is given in Sec.~\ref{sec:apx:proof_epkl_mi_rmi} in the appendix.
In Sec.~\ref{sec:old_measures} we concluded that the core problem of
the current measure of predictive uncertainty is,
that it assumes the BMA predictive distribution is equivalent to the predictive distribution
under the true model.
The reverse mutual information exactly accounts for this mismatch,
by measuring the KL-divergence between the predictive distributions under 
all possible models weighted by their posterior probability 
and the BMA predictive distribution.

\textbf{Related Work. }
We discuss other potential measures of uncertainty, not necessarily grounded in
information theory in Sec.~\ref{sec:related_work} in the appendix.

\section{Experiments} \label{sec:experiments}

\textbf{Illustrative Example. }
First, we follow Wimmer et al. \cite{Wimmer:23} and investigate an illustrative
example of different posterior distributions of the parameter $\theta$ of a 
Bernoulli distribution.
More details and results are given in section~\ref{sec:bernoulli} in the appendix.
The results indicate, that the new measure (Eq.~\eqref{eq:expected_uncertainty})
behaves more meaningfully than the current measure (Eq.~\eqref{eq:uncertainty_mi_2}).

\textbf{ImageNet. }
Second, we investigated the common tasks of out-of-distribution (OOD) detection and adversarial example detection, %
using the predictive uncertainty
as a scoring function \citep{Ovadia:19, Filos:19, Band:21, Mukhoti:23, Schweighofer:23}.
This large-scale experiment was conducted on ImageNet-1K \citep{Deng:09},
using ImageNet-O \citep{Hendrycks:21} for OOD detection and
ImageNet-A \citep{Hendrycks:21} for adversarial example detection.
This experiment utilizes versions of the EfficientNet models \citep{Tan:19}.
We compare cyclical Stochastic Gradient Hamiltonian Monte Carlo 
(cSG-HMC) \cite{Zhang:20}, Monte Carlo Dropout (MCD) \citep{Gal:16}
and Deep Ensembles \cite{Lakshminarayanan:17}, ensembling only the last layer (DE (LL)),
as well as ensembling over different pre-trained models (DE (all)).
Details and further experiments are given in 
Sec.~\ref{sec:imagenet_experiments} in the appendix.
The results are given in Tab.~\ref{tab:res:imagenet}.

The results show, that using the new measure of (total) predictive uncertainty  
(Eq.~\eqref{eq:expected_uncertainty}) is superior to using the 
current measure (Eq.~\eqref{eq:uncertainty_mi_2})
for three out of four methods, and equally good for the fourth method. 
Also, using the epistemic component of the new measure 
(Eq.~\eqref{eq:expected_uncertainty}) is, 
except for DE (all) on ImageNet-A, superior to using the 
epistemic component of the current measure (Eq.~\eqref{eq:uncertainty_mi_2}).

\setlength{\tabcolsep}{9pt} %
\renewcommand{\arraystretch}{1.3}
\begin{table}
\centering
\caption[]{AUROC using uncertainty measures as score to distinguish in-distribution (ImageNet-1K) and out-of-distribution samples (ImageNet-O) as well as natural adversarial examples (ImageNet-A). \\
\vspace{-0.cm}
}
\label{tab:res:imagenet}
\begin{tabular}{ccccccc}
\hline
\textbf{Task}               & \textbf{Method} & \multicolumn{2}{c}{\textbf{Total}}   & \textbf{Aleatoric} & \multicolumn{2}{c}{\textbf{Epistemic}} \\  \cline{3-4} \cline{6-7}
                            &            & Eq.~\eqref{eq:uncertainty_mi_2}                 & Eq.~\eqref{eq:expected_uncertainty}             &             & Eq.~\eqref{eq:uncertainty_mi_2}                    & Eq.~\eqref{eq:expected_uncertainty} \\ \hline
                            
\multirow{4}{*}{\rotatebox[origin=c]{90}{ImageNet-O}} 
                            & cSG-HMC         & $.609_{\pm .007}$ & $\boldsymbol{.611}_{\pm .007}$ & $.606_{\pm .007}$ & $.675_{\pm .006}$ & $\boldsymbol{.689}_{\pm .007}$ \\
                            & MCD             & $.631_{\pm .004}$ & $\boldsymbol{.633}_{\pm .004}$ & $.629_{\pm .004}$ & $.682_{\pm .008}$ & $\boldsymbol{.720}_{\pm .006}$ \\
                            & DE (LL)         & $\boldsymbol{.600}_{\pm .005}$ & $\boldsymbol{.600}_{\pm .005}$ & $.600_{\pm .005}$ & $.561_{\pm .002}$ & $\boldsymbol{.605}_{\pm .008}$ \\
                            & DE (all)        & $.703_{\pm .004}$ & $\boldsymbol{.709}_{\pm .005}$ & $.696_{\pm .004}$ & $.717_{\pm .007}$ & $\boldsymbol{.718}_{\pm .008}$ \\ \hline
                            
\multirow{4}{*}{\rotatebox[origin=c]{90}{ImageNet-A}}
                            & cSG-HMC         & $.677_{\pm .001}$ & $\boldsymbol{.687}_{\pm .001}$ & $.666_{\pm .001}$ & $.785_{\pm .000}$ & $\boldsymbol{.792}_{\pm .000}$ \\
                            & MCD             & $.795_{\pm .002}$ & $\boldsymbol{.797}_{\pm .002}$ & $.794_{\pm .002}$ & $.829_{\pm .001}$ & $\boldsymbol{.860}_{\pm .001}$ \\
                            & DE (LL)         & $\boldsymbol{.819}_{\pm .003}$ & $\boldsymbol{.819}_{\pm .003}$ & $.818_{\pm .003}$ & $.694_{\pm .003}$ & $\boldsymbol{.813}_{\pm .002}$ \\
                            & DE (all)        & $.887_{\pm .002}$ & $\boldsymbol{.892}_{\pm .002}$ & $.879_{\pm .002}$ & $\boldsymbol{.890}_{\pm .002}$ & $.880_{\pm .002}$ \\ \hline
\end{tabular}
\vspace{-0.cm}
\end{table}
\renewcommand{\arraystretch}{1}

\section{Conclusion and Future Work}
We analyzed the limitations of current measures of predictive uncertainty,
giving the new insight that their deficiency comes from assuming
the BMA predictive distribution is the true predictive distribution.
Therefore, we introduced a principled measure of predictive uncertainty
that address those limitations.
We showed that this new set of measures exhibits more sensible behavior and 
improves performance on common tasks where measures of predictive uncertainty are utilized.
Future work should shed light on how those measures compare in active learning
settings, where epistemic uncertainty is used to select the most informative
datapoints.

\clearpage

\section*{Acknowledgements}

The ELLIS Unit Linz, the LIT AI Lab, the Institute for Machine Learning, are supported by the Federal State Upper Austria. We thank the projects AI-MOTION (LIT-2018-6-YOU-212), DeepFlood (LIT-2019-8-YOU-213), Medical Cognitive Computing Center (MC3), INCONTROL-RL (FFG-881064), PRIMAL (FFG-873979), S3AI (FFG-872172), DL for GranularFlow (FFG-871302), EPILEPSIA (FFG-892171), AIRI FG 9-N (FWF-36284, FWF-36235), AI4GreenHeatingGrids(FFG- 899943), INTEGRATE (FFG-892418), ELISE (H2020-ICT-2019-3 ID: 951847), Stars4Waters (HORIZON-CL6-2021-CLIMATE-01-01). We thank Audi.JKU Deep Learning Center, TGW LOGISTICS GROUP GMBH, Silicon Austria Labs (SAL), FILL Gesellschaft mbH, Anyline GmbH, Google, ZF Friedrichshafen AG, Robert Bosch GmbH, UCB Biopharma SRL, Merck Healthcare KGaA, Verbund AG, GLS (Univ. Waterloo) Software Competence Center Hagenberg GmbH, T\"{U}V Austria, Frauscher Sensonic, TRUMPF and the NVIDIA Corporation.

\bibliography{arxiv}
\bibliographystyle{plain}

\clearpage
\appendix
\section{Theoretical Results}

\subsection{Equivalence of Eq.~(1) and Eq.~(2)} \label{sec:equivalence_mi}

We want to show that Eq.~\eqref{eq:uncertainty_mi_1} and Eq.~\eqref{eq:uncertainty_mi_2}
are equivalent.
The aleatoric component is already the same for both.
Therefore, we need to show that the total components are equivalent:
\begin{align}
    \rH ( p(\By \mid \Bx, \cD ) ) &=  \dE_{p(\By \mid \Bx, \cD )} \left[ - \log p(\By \mid \Bx, \cD) \right] \\
    &= \dE_{p(\Bw \mid \cD )} \left[ \dE_{p(\By \mid \Bx, \Bw )} \left[ - \log p(\By \mid \Bx, \cD) \right] \right] \\
    &= \dE_{p(\tilde\Bw \mid \cD)} \left[ \mathrm{CE} (p(\By \mid \Bx, \Bw ) \ , \ p(\By \mid \Bx, \cD )) \right]
\end{align}
Furthermore, we need to show that the epistemic components are equivalent:
\begin{align}
    \rI(p(\By, \Bw \mid \Bx, \cD)) &= \dE_{p(\By, \Bw \mid \Bx, \cD)} \left[ \log \frac{p(\By, \Bw \mid \Bx, \cD)}{p(\By \mid \Bx, \cD) \ p(\Bw \mid \Bx, \cD)} \right] \\
    &= \dE_{p(\By, \Bw \mid \Bx, \cD)} \left[ \log \frac{p(\By \mid \Bx, \Bw) \ p(\Bw \mid \cD)}{p(\By \mid \Bx, \cD) \ p(\Bw \mid \cD)} \right] \\
    &= \dE_{p(\By, \Bw \mid \Bx, \cD)} \left[ \log \frac{p(\By \mid \Bx, \Bw)}{p(\By \mid \Bx, \cD)} \right] \\
    &= \dE_{p(\Bw \mid \cD)} \left[ \dE_{p(\By \mid \Bx, \Bw)} \left[ \log \frac{p(\By \mid \Bx, \Bw)}{p(\By \mid \Bx, \cD)} \right] \right] \\
    &= \dE_{p(\Bw \mid \cD )} \left[ \mathrm{D}_{\mathrm{KL}}(p(\By \mid \Bx, \Bw ) \mid\mid p(\By \mid \Bx, \cD )) \right]
\end{align}
This holds, as $\By$ depends on $\Bw$, i.e. the model $\Bw$ has to be selected 
before being able to draw $\By$ from its predictive distribution.
Consequently, Eq.~\eqref{eq:uncertainty_mi_1} and Eq.~\eqref{eq:uncertainty_mi_2}
are equivalent.\\
\qed

\subsection{Proof of Additive Decomposition of Expected Pairwise KL-Divergence into Mutual Information and Reverse Mutual Information} \label{sec:apx:proof_epkl_mi_rmi}

Given are the definitions of the expected pairwise KL-divergence
\begin{align} \label{eq:def:epkl}
    \rK(p(\By, \Bw \mid \Bx, \cD)) \ = \ \dE_{p(\Bw \mid \cD)} \left[\dE_{p(\tilde\Bw \mid \cD)} \left[ \mathrm{D}_{\mathrm{KL}} ( p(\By \mid \Bx, \Bw) \mid\mid p(\By \mid \Bx, \tilde\Bw) ) \right] \right] \ ,
\end{align}
the mutual information
\begin{align}  \label{eq:def:mi}
    \rI(p(\By, \Bw \mid \Bx, \cD)) \ = \ \dE_{p(\Bw \mid \cD)} \left[ \mathrm{D}_{\mathrm{KL}} ( p(\By \mid \Bx, \Bw) \mid\mid p(\By \mid \Bx, \cD) ) \right] \ ,
\end{align}
and the reverse mutual information
\begin{align}  \label{eq:def:rmi}
    \rM(p(\By, \Bw \mid \Bx, \cD)) \ = \ \dE_{p(\tilde{\Bw} \mid \cD )} \left[ \mathrm{D}_{\mathrm{KL}}(p(\By \mid \Bx, \cD ) \mid\mid p(\By \mid \Bx, \tilde{\Bw} )) \right] \ .
\end{align}
We want to show that
\begin{align}
    \rK(p(\By, \Bw \mid \Bx, \cD)) = \rI(p(\By, \Bw \mid \Bx, \cD)) \ + \ \rM(p(\By, \Bw \mid \Bx, \cD)) \ .
\end{align}
Note that $p(\tilde\Bw \mid \cD) = p(\Bw \mid \cD)$ is used redundantly to keep track of integration variables.
The proof is as follows.

\begin{align} \label{eq:derivation_epkl_mi_rmi}
    \rK (p&(\By, \Bw \mid \Bx, \cD)) \\ 
    = \ &\rI(p(\By, \Bw \mid \Bx, \cD)) \ + \ \rM(p(\By, \Bw \mid \Bx, \cD)) \\
    \label{eq:derivation_epkl_mi_rmi:step_1}
    = \ &\dE_{p(\Bw \mid \cD)} \left[ \mathrm{D}_{\mathrm{KL}} ( p(\By \mid \Bx, \Bw) \mid\mid p(\By \mid \Bx, \cD) ) \right] \ + \  \dE_{p(\tilde{\Bw} \mid \cD )} \left[ \mathrm{D}_{\mathrm{KL}}(p(\By \mid \Bx, \cD ) \mid\mid p(\By \mid \Bx, \tilde{\Bw} )) \right] \\
    \label{eq:derivation_epkl_mi_rmi:step_2}
    = \ &\dE_{p(\Bw \mid \cD)} \left[ \dE_{p(\By \mid \Bx, \Bw)} \left[ \log \frac{p(\By \mid \Bx, \Bw)}{p(\By \mid \Bx, \cD)} \right] \right] \ + \ \dE_{p(\tilde\Bw \mid \cD)} \left[ \dE_{p(\By \mid \Bx, \cD)} \left[ \log \frac{p(\By \mid \Bx, \cD)}{p(\By \mid \Bx, \tilde\Bw)} \right] \right] \\
    \label{eq:derivation_epkl_mi_rmi:step_3}
    = \ &\dE_{p(\Bw \mid \cD)} \left[ \dE_{p(\By \mid \Bx, \Bw)} \left[ \log p(\By \mid \Bx, \Bw) \right] \ - \ \dE_{p(\By \mid \Bx, \Bw)} \left[ \log {p(\By \mid \Bx, \cD)} \right] \right] \ + \\ \nonumber
    &\dE_{p(\tilde\Bw \mid \cD)} \left[ \dE_{p(\By \mid \Bx, \cD)} \left[ \log {p(\By \mid \Bx, \cD)} \right] \ - \ \dE_{p(\By \mid \Bx, \cD)} \left[ \log p(\By \mid \Bx, \tilde\Bw) \right] \right] \\
    \label{eq:derivation_epkl_mi_rmi:step_4}
    = \ &\dE_{p(\Bw \mid \cD)} \left[ \dE_{p(\By \mid \Bx, \Bw)} \left[ \log p(\By \mid \Bx, \Bw) \right] \right] \ - \ \dE_{p(\By \mid \Bx, \cD)} \left[ \log {p(\By \mid \Bx, \cD)} \right] \ + \\ \nonumber
    &\dE_{p(\By \mid \Bx, \cD)} \left[ \log {p(\By \mid \Bx, \cD)} \right] \ - \ \dE_{p(\tilde\Bw \mid \cD)} \left[ \dE_{p(\By \mid \Bx, \cD)} \left[ \log p(\By \mid \Bx, \tilde\Bw) \right] \right] \\
    \label{eq:derivation_epkl_mi_rmi:step_5}
    = \ &\dE_{p(\Bw \mid \cD)} \left[ \dE_{p(\By \mid \Bx, \Bw)} \left[ \log p(\By \mid \Bx, \Bw) \right] \right] \ - \\ \nonumber 
    &\dE_{p(\tilde\Bw \mid \cD)} \left[ \dE_{p(\Bw \mid \cD)} \left[ \dE_{p(\By \mid \Bx, \Bw)} \left[ \log p(\By \mid \Bx, \tilde\Bw) \right] \right] \right] \\
    \label{eq:derivation_epkl_mi_rmi:step_6}
    = \ &\dE_{p(\Bw \mid \cD)} \left[ \dE_{p(\tilde\Bw \mid \cD)} \left[ \dE_{p(\By \mid \Bx, \Bw)} \left[ \log p(\By \mid \Bx, \Bw) \right] \right] \right] \ - \\ \nonumber 
    &\dE_{p(\Bw \mid \cD)} \left[ \dE_{p(\tilde\Bw \mid \cD)} \left[ \dE_{p(\By \mid \Bx, \Bw)} \left[ \log p(\By \mid \Bx, \tilde\Bw) \right] \right] \right] \\
    \label{eq:derivation_epkl_mi_rmi:step_7}
    = \ &\dE_{p(\Bw \mid \cD)} \left[ \dE_{p(\tilde\Bw \mid \cD)} \left[ \dE_{p(\By \mid \Bx, \Bw)} \left[ \log \frac{p(\By \mid \Bx, \Bw)}{p(\By \mid \Bx, \tilde\Bw)} \right] \right] \right] \\
    \label{eq:derivation_epkl_mi_rmi:step_8}
    = \ &\dE_{p(\Bw \mid \cD)} \left[\dE_{p(\tilde\Bw \mid \cD)} \left[ \mathrm{D}_{\mathrm{KL}} ( p(\By \mid \Bx, \Bw) \mid\mid p(\By \mid \Bx, \tilde\Bw) ) \right] \right] \ ,
\end{align}

which is exactly the definition of $\rK$ in \eqref{eq:def:epkl}.
The step from \eqref{eq:derivation_epkl_mi_rmi:step_3} to \eqref{eq:derivation_epkl_mi_rmi:step_4} is due to additivity and linearity of expectations.
The step from \eqref{eq:derivation_epkl_mi_rmi:step_5} to \eqref{eq:derivation_epkl_mi_rmi:step_6} is due to the fact that we can insert the expectation $\dE_{p(\tilde\Bw \mid \cD)}$ in the first term as it does not depend on $\tilde\Bw$ and due to the fact that $p(\tilde\Bw \mid \cD) = p(\Bw \mid \cD)$.\\
\qed

\subsection{Overview of Measures of Uncertainty} \label{sec:apx:overview_measures}

In the following, we give an overview of the measures of uncertainty
discussed in the main paper.
As the aleatoric components are equivalent, we
categorize them by the epistemic component.
The current measure of predictive uncertainty, where the epistemic component is the mutual information is given by
\begin{align} \label{eq:apx:mi}
    & \color[HTML]{D62728} \underbrace{\dE_{p(\Bw \mid \cD)} \left[ \mathrm{CE} (p(\By \mid \Bx, \Bw ) \ , \ p(\By \mid \Bx, \cD )) \right]}_{\text{total}} \\ \nonumber 
    & \qquad\qquad = \color[HTML]{D62728} \ \underbrace{\dE_{p(\Bw \mid \cD )} \left[ \rH  (p(\By \mid \Bx, \Bw ) ) \right]}_{\text{aleatoric}} \color{black} \ + \ \color[HTML]{D62728} \underbrace{\dE_{p(\Bw \mid \cD )} \left[ \mathrm{D}_{\mathrm{KL}}(p(\By \mid \Bx, \Bw ) \mid\mid p(\By \mid \Bx, \cD )) \right]}_{\text{epistemic}} \ .
\end{align}
Our new measure of predictive uncertainty, where the epistemic component is the expected pairwise KL-divergence is given by
\begin{align} \label{eq:apx:epkl}
     & \underbrace{\dE_{p(\Bw \mid \cD)} \left[ \dE_{p(\tilde\Bw \mid \cD)} \left[ \mathrm{CE} ( p(\By \mid \Bx, \Bw) \ , \ p(\By \mid \Bx, \tilde\Bw) ) \right] \right]}_{\text{total}} \\ \nonumber
    & \qquad = \ \underbrace{\dE_{p(\Bw \mid \cD)} \left[\rH  (p(\By \mid \Bx, \Bw ) ) \right]}_{\text{aleatoric}} \ + \     \underbrace{\dE_{p(\Bw \mid \cD)} \left[\dE_{p(\tilde\Bw \mid \cD)} \left[ \mathrm{D}_{\mathrm{KL}} ( p(\By \mid \Bx, \Bw) \mid\mid p(\By \mid \Bx, \tilde\Bw) ) \right] \right]}_{\text{epistemic}} \ .
\end{align}
Furthermore, it is possible to decompose our new measure of predictive uncertainty,
such that the reverse mutual information is the epistemic component:
\begin{align} \label{eq:apx:rmi}
     & \underbrace{\dE_{p(\Bw \mid \cD)} \left[ \dE_{p(\tilde\Bw \mid \cD)}  \left[ \mathrm{CE} (p(\By \mid \Bx, \Bw) \ , \ p(\By \mid \Bx, \tilde\Bw )) \right] \right]}_{\text{total (Eq.~\eqref{eq:apx:epkl})}} \\ \nonumber \color{black}
    & \qquad\qquad = \color[HTML]{D62728}\ \underbrace{\dE_{p(\Bw \mid \cD)} \left[ \mathrm{CE} (p(\By \mid \Bx, \Bw ) \ , \ p(\By \mid \Bx, \cD )) \right]}_{\text{total (Eq.~\eqref{eq:apx:mi})}} \color{black} \ + \ \color[HTML]{9467BD}\underbrace{\dE_{p(\tilde{\Bw} \mid \cD )} \left[ \mathrm{D}_{\mathrm{KL}}(p(\By \mid \Bx, \cD ) \mid\mid p(\By \mid \Bx, \tilde{\Bw} ) ) \right]}_{\text{epistemic}} \ .
\end{align}
However, this does not yield an exclusively aleatoric component, but the 
(total) predictive uncertainty of Eq.~\eqref{eq:apx:mi}.
This is also signified by the coloring of the respective terms in Eq.~\eqref{eq:apx:rmi}.

\section{Experimental Results}

\subsection{Illustrative Example: Bernoulli Distribution}\label{sec:bernoulli}

\paragraph{Setup.}
Following Wimmer et al. \citep{Wimmer:23}, we analyze the behaviour of
different measures of uncertainty using a posterior distribution
over the distribution parameter $\theta \in [0, 1]$ of a Bernoulli distribution.
In this setting, \emph{no model is involved}, thus the distribution parameter $\theta$
is directly sampled from the posterior distribution.
Furthermore, we drop the dependence on a specific input $\Bx$,
assuming that the input gives no information about the outcome.
For the sake of an example, consider a machine where one can
press a button and one of two signals, red (0) or green (1) will light up.
The task is to predict which signal will light up.

We analyze the behavior of the measures of uncertainty under several 
closed-form posterior distributions, such as uniform, beta 
and a mixture of delta distributions.
Those correspond to different knowledge about which signal will light up.
For example, a delta distribution at $\theta = 0$ would correspond to that
it is known the outcome is either green or red with equal probability.
Thus the outcome is maximally uncertain, due to aleatoric uncertainty.
A different case would be a uniform posterior distribution, thus every
parameter $\theta \in [0, 1]$ is equally probable.
There should be less aleatoric uncertainty, as parameters that
lead to Bernoulli distributions certain in their prediction are also possible.
However, epistemic uncertainty must be higher,
as all parameters are equally probable.
Where should we expect the highest epistemic uncertainty?
One point could be made that we should expect the highest
epistemic uncertainty under the uniform distribution,
due to the principle of maximum entropy \citep{Jaynes:57}.
Therefore, if all parameters are equally probable, epistemic uncertainty should be highest.
However, this captures uncertainty in the parameter space, not in the
output space, which is what we are concerned about for predictive uncertainty.
We don't care, how many models are possible, but about how different their predictions are.
Therefore, we argue that a mixture of two delta distributions at the two extreme parameters
$\theta = 0$ and $\theta = 1$ should correspond to the highest epistemic uncertainty
in this example.

How could such a situation occur?
This reflects the prior belief that the machine in our example is deterministic in nature,
thus no matter how often the button is pressed, the same signal will light up.
However, before trying it once, it is unknown which signal will light up.
Therefore, when choosing the parameter corresponding to the green light,
there would be maximal surprisal when observing the red light.
Yet after observing a single outcome, the posterior will update,
collapsing to a delta distribution at a single parameter,
removing all uncertainty about the prediction.
Still, before observing a single outcome, 
the epistemic uncertainty should be maximal.

\begin{figure}
    \centering
    \includegraphics[width=\textwidth]{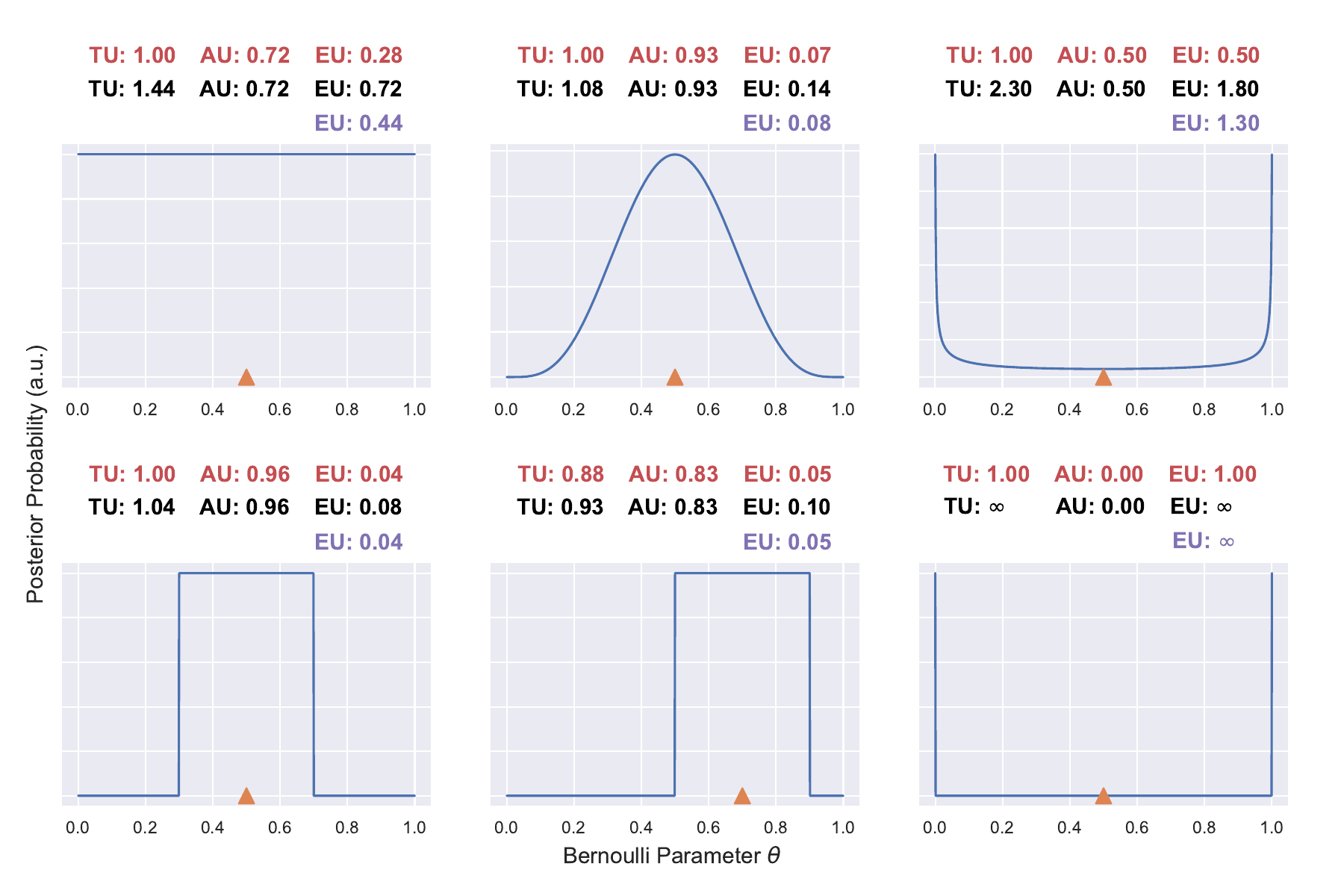}
    \caption{Different posterior distributions $p(\theta \mid \cD)$ for the parameter $\theta$ of a Bernoulli distribution. Uncertainties using the current measure of predictive uncertainty (Eq.~\eqref{eq:uncertainty_mi_2}), the introduced measure of predictive uncertainty (Eq.~\eqref{eq:expected_uncertainty}), and the reverse mutual information as measure for epistemic uncertainty (Eq.~\eqref{eq:apx:rmi}). Orange triangles denote the expected value of $\theta$ under the posterior distribution. Posterior distributions from left to right, per row: $\cU\left[0, 1\right]$, $Beta(5, 5)$, $Beta(0.4, 0.4)$,  $\cU\left[0.3, 0.7\right]$, $\cU\left[0.5, 0.9\right]$, $\frac{1}{2}\delta_0 + \frac{1}{2}\delta_1$.}
    \label{fig:bernoulli}
\end{figure}

\paragraph{Results.}
Different posterior distributions and associated total (TU), 
aleatoric (AU) and epistemic uncertainty (EU)
under the considered measures of uncertainty are depicted in Fig.~\ref{fig:bernoulli}.

For the current measure of uncertainty given by Eq.~\eqref{eq:uncertainty_mi_2},
results are given in the first row above each plot, written in red.
We find that total uncertainty is maximal if the Bernoulli parameter 
of the BMA is $\theta = 0.5$.
This does not depend on how the probability mass of the posterior is distributed,
but only on the expectation.
For the introduced measure of uncertainty given by Eq.~\eqref{eq:expected_uncertainty},
results are given in the second row above each plot, written in black.
We find, that the measure for epistemic uncertainty gives infinity,
and consequently also the measure for total uncertainty.
This corresponds exactly to the expected behavior of an information-theoretic measure
of uncertainty about the outcome.
We also consider the reverse mutual information as a standalone measure of epistemic uncertainty \citep{Malinin:21}.
Results are given in the third row above each plot, written in violet.
We find, that the behavior of this measure of epistemic uncertainty is
similar to the epistemic component of the measure given by Eq.~\eqref{eq:expected_uncertainty}.

\begin{figure}
    \centering
    \includegraphics[width=\textwidth]{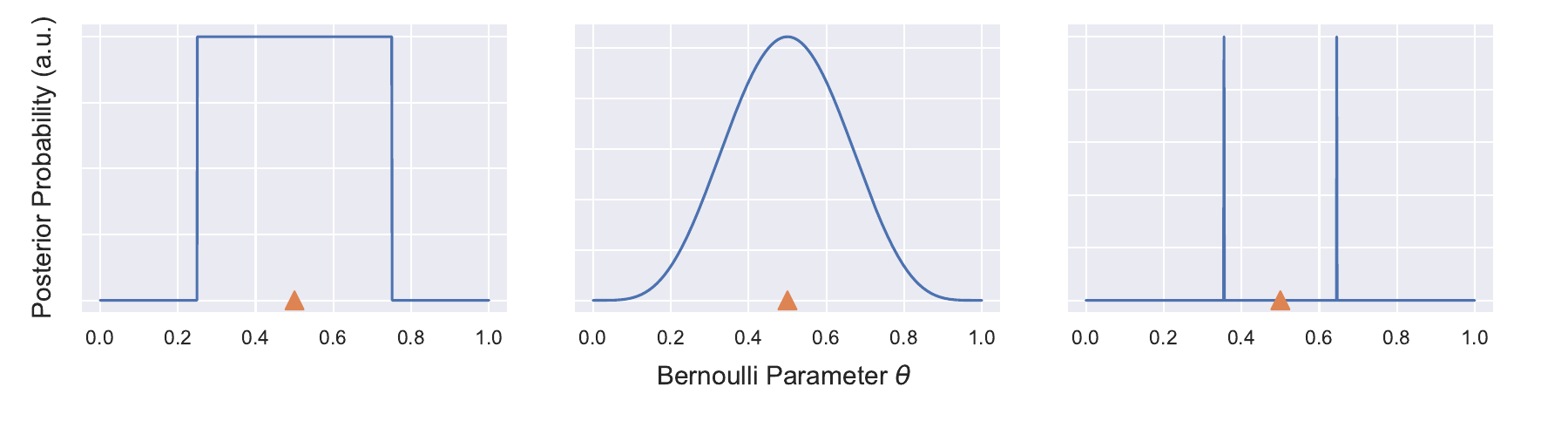}
    \caption{Three different posterior distributions $p(\theta \mid \cD)$ (Uniform, Beta, Mixture of delta distributions) for the parameter $\theta$ of a Bernoulli distribution. All of them have the same total, aleatoric, and epistemic uncertainty according to Eq.~\eqref{eq:uncertainty_mi_2}. Those can easily be found, as the total uncertainty only depends on the expected value of $\theta$ under the posterior distribution, depicted by the orange triangle. Fixing the expected value of the posterior distribution, the distribution parameters are trivial to tune to exhibit the same aleatoric and consequently the same epistemic uncertainty. However, those posterior distributions exhibit different total and epistemic uncertainty according to Eq.~\eqref{eq:expected_uncertainty}.}
    \label{fig:mi_problem}
\end{figure}

\paragraph{Another Limitation of Current Measure of Uncertainty.}
Another limitation of the current measure of uncertainty given by Eq.~\eqref{eq:uncertainty_mi_2}
is exemplified in Fig.~\ref{fig:mi_problem}.
Here, all three posterior distributions lead to equal total, aleatoric, and epistemic uncertainty.
Such examples are easy to construct.
The total uncertainty given by Eq.~\eqref{eq:uncertainty_mi_2} 
(better seen in the equivalent formulation in Eq.~\eqref{eq:uncertainty_mi_1}) 
only depends on the expected value (the BMA) of the Bernoulli parameter $\theta$.
By fixing the expected value to e.g. $\theta = 0.5$, it is trivial to search for distribution
parameters that lead to equal aleatoric and, due to additivity, also to equal epistemic
uncertainty.
Using the measure of uncertainty given by Eq.~\eqref{eq:uncertainty_mi_2}
resolves the ambiguity, assigning different epistemic and therefore also total
uncertainty to the three posterior distributions.

\clearpage
\subsection{ImageNet Experiments} \label{sec:imagenet_experiments}

ImageNet experiments were conducted using the codebase of 
\citep{Schweighofer:23} with their default hyperparameters. 

Additionally to the out-of-distribution (OOD) detection and adversarial example detection tasks 
reported in Tab.~\ref{tab:res:imagenet} of the main paper, 
we investigated the common tasks of misclassification detection and selective prediction, again 
using the predictive uncertainty as a scoring function \citep{Ovadia:19, Filos:19, Band:21, Mukhoti:23, Schweighofer:23}. 
We conducted those experiments on the official validation set of ImageNet-1K. 

For each of the four experiments, we use pre-trained EfficientNet \citep{Tan:19} architectures available through PyTorch \citep{Paszke:19}. 
To approximate the posterior expectations in Eq.~\eqref{eq:uncertainty_mi_2} and Eq.~\eqref{eq:expected_uncertainty}, we utilize the standard approach of MC integration thus approximating the integral by the average over functions of model parameters (approximately) drawn from the posterior distribution.
As an example, the BMA predictive distribution is approximated by 
\begin{align}
p(\By \mid \Bx, \cD) = \dE_{p(\Bw \mid \cD)} \left[ p(\By \mid \Bx, \Bw) \right] \approx \frac{1}{N} \sum_{i=1}^N p(\By \mid \Bx, \Bw_i) \ ,
\end{align}
where $\Bw_i \sim p(\Bw \mid \cD)$.
To sample different model parameters, we utilize cSG-HMC \citep{Zhang:20}, MC dropout \citep{Gal:16} and versions of Deep Ensembles \cite{Lakshminarayanan:17}.
cSG-HMC and MC dropout were performed on the last layer of the EfficientNetV2-S architecture with 2000 samples each. 
Deep Ensembles were used in two different versions.
Once by ensembling 10 different pre-trained EfficientNet models with different architecture (DE (all)) and once by ensembling 10 different last layers of the EfficientNetV2-S architecture (DE (LL)).

To provide confidence intervals, we performed all experiments on three distinct dataset splits. 
Regarding OOD detection, each split consists of all 2000 Imagenet-O samples and 2000 unique ImageNet-1K samples. 
Regarding adversarial example detection, each split consists of all 7000 Imagenet-A samples and 7000 unique ImageNet-1K samples. 
Regarding misclassification detection and selective prediction, each spit consists of 7000 unique ImageNet-1K samples. 
ImageNet-1K samples were randomly drawn from the official validation set.

The additional results for misclassification detection and selective prediction are given in Tab.~\ref{tab:imagenet_a_o_results}.
The results show, that using the new measure of (total) predictive uncertainty 
given by Eq.~\eqref{eq:expected_uncertainty},
is always at least equally good and most of the time better then using
the current measure of (total) predictive uncertainty given by Eq.~\eqref{eq:uncertainty_mi_2}.
The only exception is DE (all) for both tasks.
Results for the epistemic components of both measures are similar.

\setlength{\tabcolsep}{9pt} %
\renewcommand{\arraystretch}{1.3}
\begin{table}
\centering
\caption[]{AUROC using different uncertainty measures as a score to distinguish between correctly and misclassified samples on the ImageNet-1K validation set. For selective prediction, the AUC of accuracy vs. fraction of most certain samples is reported.\\
\vspace{-0.2cm}
}
\label{tab:imagenet_a_o_results}
\begin{tabular}{ccccccc}
\hline
\textbf{Task}               & \textbf{Method} & \multicolumn{2}{c}{\textbf{Total}}   & \textbf{Aleatoric} & \multicolumn{2}{c}{\textbf{Epistemic}} \\  \cline{3-4} \cline{6-7}
                            &            & Eq.~\eqref{eq:uncertainty_mi_2}                 & Eq.~\eqref{eq:expected_uncertainty}             &             & Eq.~\eqref{eq:uncertainty_mi_2}                    & Eq.~\eqref{eq:expected_uncertainty} \\ \hline
\multirow{4}{*}{\rotatebox[origin=c]{90}{Misclass.}}
                            & cSG-HMC         & $.700_{\pm .003}$ & $\boldsymbol{.705}_{\pm .003}$ & $.693_{\pm .003}$ & $.758_{\pm .010}$ & $\boldsymbol{.767}_{\pm .009}$ \\
                            & MCD             & $\boldsymbol{.867}_{\pm .007}$ & $\boldsymbol{.867}_{\pm .007}$ & $.866_{\pm .007}$ & $.791_{\pm .012}$ & $\boldsymbol{.831}_{\pm .011}$ \\
                            & DE (LL)         & $\boldsymbol{.910}_{\pm .007}$ & $\boldsymbol{.910}_{\pm .007}$ & $.910_{\pm .007}$ & $.661_{\pm .002}$ & $\boldsymbol{.824}_{\pm .006}$ \\
                            & DE (all)        & $\boldsymbol{.879}_{\pm .004}$ & $.869_{\pm .005}$ & $.883_{\pm .004}$ & $\boldsymbol{.808}_{\pm .011}$ & $.795_{\pm .013}$ \\ \hline
                                                        
\multirow{4}{*}{\rotatebox[origin=c]{90}{Select. Pred.}}
                            & cSG-HMC         & $.911_{\pm .005}$ & $\boldsymbol{.913}_{\pm .005}$ & $.910_{\pm .005}$ & $\boldsymbol{.926}_{\pm .007}$ & $\boldsymbol{.926}_{\pm .006}$ \\
                            & MCD             & $\boldsymbol{.959}_{\pm .003}$ & $\boldsymbol{.959}_{\pm .003}$ & $.958_{\pm .003}$ & $.934_{\pm .009}$ & $\boldsymbol{.945}_{\pm .008}$ \\
                            & DE (LL)        & $\boldsymbol{.971}_{\pm .002}$  & $\boldsymbol{.971}_{\pm .002}$ & $.971_{\pm .002}$ & $.892_{\pm .005}$ & $\boldsymbol{.926}_{\pm .005}$ \\
                            & DE (all)        & $\boldsymbol{.968}_{\pm .001}$ & $.966_{\pm .001}$ & $.969_{\pm .001}$ & $\boldsymbol{.953}_{\pm .003}$ & $.950_{\pm .003}$ \\ \hline
                            
\end{tabular}
\end{table}
\renewcommand{\arraystretch}{1}

\clearpage
\section{Related Work} \label{sec:related_work}

Mutual information as a measure of epistemic uncertainty (about the parameters of the model) 
and the respective decomposition into total and aleatoric uncertainty
was introduced by Houlsby et al. \citep{Houlsby:11}.
Those measures have remained popular ever since \citep{Gal:16thesis, Smith:18, Depeweg:18, Huellermeier:21, Mukhoti:23}.
Smith et al. \citep{Smith:18} analyzed measures of uncertainty for adversarial example detection.
Furthermore, they illustrate the relation of mutual information to another ad-hoc measure
for uncertainty in classification settings, the softmax variance \citep{Leibig:17, Carlini:17}.
Another ad-hoc measure of uncertainty in classification settings is the maximum softmax value
of a classifier \citep{Lakshminarayanan:17, Hendrycks:18, Gal:16thesis}.

Depeweg et al. \citep{Depeweg:18} considers the measures in Eq.~\eqref{eq:uncertainty_mi_1},
but also proposes to use the variance as a measure of uncertainty.
Based upon the law of total variance, the variance of the BMA predictive distribution
is decomposed to a posterior expectation over the variance of the predictive distributions given by individual models
(the aleatoric component) and the posterior variance over the expected value of predictive distributions given by
individual models (the epistemic component).

Malinin et al. \citep{Malinin:18} introduces measures similar to Eq.~\eqref{eq:uncertainty_mi_1},
but modeling the mutual information between the prediction and the data distribution.
This considers epistemic uncertainty arising from distributional mismatch between the data 
distribution during training and inference.
Furthermore, they consider the differential entropy of their introduced Dirichlet Prior Network,
which captures the entropy of the predicted Dirichlet distribution.

Measuring predictive uncertainty through the information-theoretic measures discussed in this work
are based upon posterior expectations, which are approximated by Monte-Carlo sampling
in practice.
This requires to sample multiple models and obtain predictions for them, making it
expensive at inference time.
Therefore, methods to estimate epistemic uncertainty with just a single pass through the
network were considered \citep{Bradshaw:17, Liu:20, VanAmersfoort:20, VanAmersfoort:22, Mukhoti:23}.
They measure epistemic uncertainty via feature-space distance or feature-space densities.
Thus, epistemic uncertainty is high if a new input is far from samples in the training set
or the density of the training set is low.
However, this requires that the feature space is meaningful, 
where feature collapse might severely impact the quality of the uncertainty estimate \citep{VanAmersfoort:22, Postels:22}.

Apart from distributional representations of uncertainty, set-based formalisms are
alternative ways to express uncertainty, e.g. using credal sets \citep{Huellermeier:21, Huellermeier:22b}.
Sale et al. \citep{Sale:23} investigated the volume of the credal set as a measure for
epistemic uncertainty, but found it is only meaningful in the case of binary classification
and less so for multiclass-classification.

\end{document}